\pdfoutput=1
\documentclass[11pt]{article}

\usepackage[preprint]{acl}

\usepackage{times}
\usepackage{latexsym}

\usepackage[T1]{fontenc}
\usepackage[utf8]{inputenc}

\usepackage{microtype}

\usepackage{inconsolata}

\usepackage{graphicx}

\usepackage{tabularx}
\usepackage{makecell}
\usepackage{adjustbox}
\usepackage{bm}

\usepackage{mathtools}
\usepackage{mathpazo}
\usepackage{multirow}

\usepackage{pifont}
\newcommand{\cmark}{\ding{51}}%
\newcommand{\xmark}{\ding{55}}%
\newcommand{\crossedcmark}{\textcolor{black}{\ding{51}}{\small\textcolor{black}{\kern-0.7em\ding{55}}}}
\usepackage{tikz,graphics,color,epsf,subcaption}

\usepackage[show]{notes}
\newcommand{\ignore}[1]{}
\newcommand{\com}[1]{}

\usepackage{xcolor}

\title{AggregHate:\\An Efficient Aggregative Approach for the Detection of Hatemongers on Social Platforms}

\author{Tom Marzea \\
    SISE\\
    Ben Gurion University\\
  \texttt{tommarz@post.bgu.ac.il} \\\And
  Abraham Israeli \\
  UMSI\\
  University of Michigan\\
  \texttt{isabrah@umich.edu} \\\And
  Oren Tsur \\
  SISE\\
  Ben Gurion University\\
  \texttt{orentsur@bgu.ac.il}
  }

\usepackage[show]{notes}

\usepackage{bibentry}
\begin{document}

\maketitle

\begin{abstract}

Automatic detection of online hate speech serves as a crucial step in the detoxification of the online discourse. Moreover, accurate classification can promote a better understanding of the proliferation of hate as a social phenomenon.  
While most prior work focus on the detection of hateful \emph{utterances}, we argue that focusing on the \emph{user} level is as important, albeit challenging. In this paper we consider a multimodal aggregative approach for the detection of hate-mongers, taking into account the potentially hateful texts, user activity, and the user network. 
We evaluate our methods on three unique datasets X (Twitter), Gab, and Parler showing that a processing a user's texts in her social context significantly improves the detection of hate mongers, compared to previously used text and graph-based methods. Our method can be then used to improve the classification of coded messages, dog-whistling, and racial gas-lighting, as well as inform intervention measures. Moreover, our approach is highly efficient even for very large datasets and networks.

\end{abstract}

\noindent{{\bf Disclaimer}: The illustrative examples in Table \ref{tab:hate_texts} may be offensive to some readers.}

\section{Introduction}
\label{sec:intro}
The rising popularity of social platforms enhanced the hateful content targeting minorities and led to the proliferation of online hate speech \cite{waseem2016hateful,laub2019hateGlobalComparison}. Accordingly, there is a growing body of research on the appearance and magnitude of hate speech on social media, in general, \cite{knuttila2011user, chandrasekharan2017you, zannettou2018gab}, and on hate speech detection, in particular, \cite{saleem2017web, waseem2016hateful, davidson2017automated}.
Hate speech is not merely an online inconvenience as shooting, bombing, stabbing, beating, and vandalism are often linked to online activity \cite{altrightpipeline,violence2019,8chan,mcilroy2019welcome, mathew2019spread,allenShooting23}.

Hate is not promoted by isolated individuals, but rather by communities that often exist within larger communities. Shifting the focus from the post level to the user level can be beneficial on various levels (pun intended): it provides a better understanding of the dynamics of hate groups and the ways hate infects individuals and communities; it could be reapplied in the post level, improving the detection of coded language, dog whistling and gas-lighting; finally, it could provide the social structure in which the impact of different intervention techniques\footnote{Intervention alternatives depend on legal regulation and platform policy and range from the removal of a specific post, the flagging of a specific post, to the temporal or indefinite suspension of an account.} could be tested \cite{thomas2023disrupting}.

Many of the challenges inherent to hate speech classification on the utterance level are discussed by \citet{Talat2018,fortuna-etal-2022-directions,velankar2022review, kim2024labelaware}, among others. 

A recent survey of the applicability of state-of-the-art LLMs highlights the shortcomings of the use of LLMs for nuanced tasks such as hate, toxicity and empathy \cite{ziems24llms4css}. 

 We provide some illustrative examples in Table \ref{tab:hate_texts}, complemented with a short analysis, demonstrating the specific challenges and outlining the possible benefits of shifting our attention from the utterance (textual) level to the speaker (user) level.

\begin{table*}[ht!]
\footnotesize
\centering
\begin{tabular}{c|m{8.5cm}|c|c|c|c|c }
 & \multicolumn{1}{c|}{Text} & DHB & JS & FTDB & GPT4 & Gemini 1.5Pro \\ \hline\hline 
1  & ``Muslim terrorists did this and obama brought them here to run our government. Gotta get rid of Obammy, Omar and all these other fucking Muslims'' & .57 & .9 & .98 & \cmark & \cmark\\ \hline
2  & ``is there a more debased creature in our modern pop-cultural cesspool than (((Amy Schumer)))? Probably not.'' & .22 & .35 & .92  &\cmark $^?$  & \cmark\\\hline
3 &   ``We must secure the existence of our people and a future for white children!'' & .03 & .36 & .02 & \cmark & \cmark $^?$ \\ \hline
4  &  ``If white people are going to continue  existing - we must support our race''  & .06 & .4 & .69 &\cmark $^?$  & \cmark $^?$  \\ \hline
5  &  ``When the youngest white generation becomes adult they will be 5\% of the world population, down from 1/3 in 1900'' & .06 & .25 & .01  &\xmark & \cmark $^?$ \\ \hline
6  & ``Did you know that most dog breeds were bred by whites? We’re awesome and they’re awesome!'' & .2 & .25 & .65  &\xmark & \xmark \\ \hline
7 &  ''White lives matter!'' & .02 & .24 & .02  &\xmark  & \xmark $^?$ \\ \hline
8 &  ``Blue lives matter!'' & .02 & .01 & .01  &\xmark & \xmark $^?$ \\\hline
\end{tabular}
\caption{Examples of hate-promoting texts. DHB: the hate-likelihood of the text assigned by deHateBERT \cite{aluru2020deep}; JS: hate prediction by Google's Jigsaw; FTDB: DistillBERT \cite{Sanh2019DistilBERTAD} fine-tuned on our datasets; GPT-4 and Gemini 1.5 Pro, using the prompt ``Which of the following texts should be classified as hate speech:''. (''Which of the following utterances may be perceived as hate speech?''). A  {\bf `?'} postscript indicates that a nuanced prediction was generated by the model (see full generated predictions in Appendix \ref{sec:generated_predictions}. }
\label{tab:hate_texts}
\end{table*}

The explicit racism of utterance \#1 is detected by all models, however, deHateBERT \cite{aluru2020deep} (DHB)  -- a BERT model fine-tuned for the detection of hate speech -- assigned a likelihood of only 0.57 for it to be classified as hate speech. 
Three of the models, ChatGPT-4 included, failed to detect the antisemitism conveyed in utterance \#2 as it requires ``external'' knowledge, namely that Schumer is of Jewish heritage and that the (((echo))) symbol is a (newly) recognized hate symbol\footnote{\url{www.adl.org/resources/hate-symbol/echo}}. The text in utterance \#3, known as the `14 words' -- ``the most popular white supremacist slogan in the world''\footnote{\url{www.adl.org/resources/hate-symbol/14-words}}. However, it may appear innocent to the uninformed (human) eye, as well as to models that did not encounter it (in context) in training. As a result, it was identified only by ChatGPT-4 and Gemini 1.5 Pro.
Posts \#4-7 are thinly veiled allusions to the ``14 words''. The fine-tuned distilBERT (FTDB)  captures \#4 and \#6 but misses the ``14 words''. Both \#7 and \#8 twist the slogan `Black Lives Batter' -- originally a protest against police brutality. While \#7 carries an explicit white-supremacist tone\footnote{\url{www.adl.org/resources/hate-symbol/white-lives-matter}}, \#8\footnote{A response to the BLM movement, suggesting that attacking policemen should be considered a hate crime.} is not recognized as hate-speech, although it is often used alongside racial slurs. Even state-of-the-art models such as ChatGPT-4 and Gemini 1.5 Pro either completely fail to detect the hate conveyed in utterances 4-8, or flag it with a low confidence. 

While all of the texts in Table \ref{tab:hate_texts} pose a challenge to computational models and humans alike, contextual information and careful aggregation can be used to achieve classification on the user level: multiple implicit posts (coded, ambiguous, dog-whistling, or gas-lighting) posted by a single user can reinforce the weak signal obtained from a single post. Similarly, association with other users or explicit posts can reinforce a weak signal of a context-less post. 

Contextual aggregation, however, is not straightforward as it depends on a number of philosophical and practical considerations stemming from the likelihood or the confidence of a model (or a human) to assign the class for a specific post. 
For example, one can argue that a single, though explicit, hateful post may not be enough to label the user as a racist or a hate-monger. However, even if a single explicit post is sufficient -- what would be the user label in case of two \emph{implicit} posts, or a few dozen of posts, each is predicted to be hateful with a low confidence? Looking at the illustrative posts in Table \ref{tab:hate_texts}, we propose a principled way to effectively combine predictions and modalities in order to achieve an accurate classification.

To this end we explore three fundamentally different approaches for contextual aggregation: (i) using binary weights with a fixed threshold, (ii) using a relational aggregation conditioned on the social context, and (iii) using sidtributional aggregation conditioned on aggregated confidence levels. Finally, we combine these methods to forn a multimodal classification model.

\paragraph{Contribution} Our contribution of this work is threefold:
\vspace{-.2cm}
\begin{enumerate}
    \item We propose a robust and efficient multimodal   aggregative approach for the detection of hate-mongers.
    \vspace{-.2cm}
    \item We demonstrate the benefits of contextual aggregation over three unique datasets (Twitter, Gab, and Parler).
    \vspace{-.2cm}
    \item We share a novel annotated dataset of Parler hate. 
\end{enumerate}
\vspace{-.2cm}

\section{Related Work}
\label{sec:related_work}
A comprehensive overview of methods and benchmarks for hate-speech detection is provided by \citet{alkomah2022}, while many of the challenges in current approaches are surveyed by \citet{elsherief2021latent,velankar2022review,fortuna-etal-2022-directions}, among others. Subjectivity and the incomplete definition of hate-speech are addressed by \citet{khurana-etal-2022-hate} while the limitations of transfer learning for the task are demonstrated by  \citet{israeli2022free}, and a set of functional tests to evaluate the performance of different models trained on different benchmarks, types, targets, and languages were proposed by \citet{rottger2020hatecheck,rottger2022multilingual}.

A taxonomy of implicit hate was developed and shared by \citet{elsherief2021latent}. Unfortunately, most of the annotated tweets are no longer available, thus user level aggregation cannot be achieved.

A growing number of works has shifted the attention from the utterance level to the user level. \citet{waseem2016hateful} analyze the relation between demographic features and hate speech, 
while \citet{ribeiro2017like} explores the differences between account meta-features of hateful and non-hateful Twitter users. Both works are of an exploratory nature. 

\citet{arviv2021sa} detects hate mongers using a multi-modal architecture that combines three streams of post-level predictions: the tweets of the target user, her followers, and her followees.

A two-step approach considering both the textual and the network modalities was proposed by \citet{ribeiro2018characterizing} and extended by \citet{israeli2022free}: In the first step 
seed nodes (users) are detected based on the textual signal (keyword matching in \citeauthor{ribeiro2018characterizing} and a fine-tuned BERT in \citeauthor{israeli2022free}). In the second step a diffusion model is applied in order to propagate the initial hate assignments across the social network.

A number of works use Graph Neural Networks (GNN) in order to detect hate speech or hate mongers. \citet{li2021neighbours} presented HateGNN, using textual similarity and the appearance of predefined hate terms as part of the objective that produces the node embeddings to be used for classification. The learned embeddings depend heavily on the training data, the type of hate, and the supervision.
\citet{miao22} proposed an end-to-end framework, enriching  a BERT classifier with Graph Attention Networks. However, in spite of the use of graph networks, this approach is applied for the detection of hate only on the post level rather than on the user level.
Several supervised, unsupervised and semi supervised models were explored by \cite{das2021you}, including the state-of-the-art GNN models such as AGNN \cite{thekumparampil2018attentionbased} in order to classify users as hateful and non-hateful.

\citet{nirmal2024interpretable} proposed \emph{SHIELD} - a framework that leverages LLM-extracted rationales to augment a base hate speech detection model to facilitate faithful interpretability. However, it was demonstrated that fine-tuned classification models outperform state-of-the-art LLMs over tasks that involve social nuances, e.g., detection of humor, empathy, toxicity and hate \cite{ziems24llms4css}.

%==============================================================================

\section{Multimodal Aggregative Approaches}
\label{sec:methods}

\subsection{Aggregative Approaches}
\label{subsec:aggregative}

\paragraph{Utterance-level Classification ($C^T$)} The basic building block of all aggregative approaches is the classification of a single utterance.
Any classification model can be used for the utterance level as long as its output can be interpreted as a probability (e.g., by applying the sigmoid function). The probability is needed in order to assign a ``confidence'' threshold, before making the binary decision. Formally:
\vspace{-.2cm}
\begin{align}
    C^T(t) = %PLM(t,M,\tau^T)=
            \begin{cases}
            1& {\theta(t)}\geq \tau^T\\
            0& \text{otherwise}
        \end{cases}
        \label{eq:plm}
\end{align}

\vspace{-.2cm}
Where $t$ is a text snippet (post, tweet, utterance), $\theta$ is a classification model, and the hyper parameter $\tau^T$ is the post-level threshold reflecting the sensitivity of the model (or the community) to implicit forms of hate speech.
In this work we follow \citet{israeli2022free} and use a DistilBERT classifier \cite{sanh2019distilbert} fine-tuned for hate-speech detection on our three datasets (see Section \ref{sec:data}).

\paragraph{User-level Classification ($C^U$)}
\label{subsec:user_level}
The detection of hateful \emph{users} is inherently related to the user's posts. Given a user $u$ and a the user level threshold $\tau^U$, the generic user classification function is given by:
\vspace{-.2cm}
\begin{align}
    C^U(u) =
            \begin{cases}
            1& {\Theta(u)\geq \tau^U}\\
            0& \text{otherwise}
        \end{cases}
        \label{eq:ulm}
\end{align}
\vspace{-.2cm}
\paragraph{Naive Aggregation with Fixed Threshold} Given $T^u$ -- the set of posts published by $u$, a naive aggregative approach could simply count the number of hateful posts (decided by the $C^T$) published by a user $u$:  
\begin{align}
    \Theta(u) = \sum_{t\in T^u}{C^T(t)}
    \label{eq:fixed}
\end{align}

While naive aggregation does not require any training, the threshold $\tau^U$ can be used to control the sensitivity of the model: setting $\tau^U=1$ implies zero-tolerance (often too harsh\footnote{Remember that in the standard case $\tau^T=0.5$ thus a user will be labeled a hate-monger even if $\theta(t)=0.51$ for one of her posts and $\theta(t')< 0.5 ~~~\forall t' \in T^u / \{t\}$.}) approach, while setting $\tau^U = a$, $a \in \{10,20, 30,...\}$ implies a more conservative approach, requiring a user to be a ``repeated offender'' in order to be labeled as a hate-monger. For convenience, we denote to the naive aggregation (fixed threshold) $\Theta_F$, and the naive classification function $C_F$.

In order to achieve flexibility and robustness one should consider other aggregation approaches that account for the nuanced and implicit utterances as well as for the social context in which a user is embedded. The different aggregative approaches could be combined to a user feature vector and $\Theta$ could be trained to optimize the weight of each feature. Specifically, instead of a naive aggregation we consider two types of aggregative functions: Relational Aggregation ($\Theta_R$) and Distributional Aggregation ($\Theta_D$). Given a an annotated dataset, we train $\Theta$ to optimize $C^U$, where $\Theta$ is a logistic regression classifier.

In the remainder of this section we motivate the use of each aggregative approach and provide the formal definition of the model.

\paragraph{Relational Aggregation} The Aristotelian concept of man as a social and political animal informed decades of sociological research concerning the importance of community to individual identity \cite{mcmillan1986sense,wellman1999virtual}. Unfortunately, identity is often forged by association with hate groups, off and on-line \cite{gordon2017second,govers2023down}. 
It is therefore reasonable to assume that hateful content circulating in $u$'s ego network should be taken into account and inform the label assigned to $u$. That is, even if a conservative $\tau^U$ is used, the associates of $u$ may push her over the threshold.

Formally, given $G(V,E)$ -- a directed social network where $V$ is the set of users and $(u,v) \in E$ indicates a directed edge $u \rightarrow v$, we define $\overleftarrow{u}$ and $\overrightarrow{u}$ as the sets of followers and followees of $u$, respectively. In the relational case, $\Theta_R(u)$ is a linear combination of three terms -- the naive aggregation over $u$'s texts and the percentages of hateful users among his followers and followees: 
\vspace{-.2cm}
\begin{multline*}
\Theta_R(u) = \alpha \cdot C_F(u) 
+ \beta \cdot \frac{1}{|\overleftarrow{u}|}\sum_{v\in{\overleftarrow{u}}} C_F(v) \\+ \gamma \cdot \frac{1}{|\overrightarrow{u}|}\sum_{v\in{\overrightarrow{u}}} C_F(v)  
\end{multline*}

\vspace{-.2cm}
The values of $\alpha$, $\beta$, and $\gamma$ are optimized through training, essentially reflecting the importance of the user's posts and the posts of his followers and followees. 
%Note that manually setting $\alpha=1$, and $\beta=\gamma=0$ reduces $\Theta_R$ to the Naive approach.

\paragraph{Distributional Aggregation} While relational aggregation takes the social context into account, it does not address variations the intensity of the promoted hate in terms of ``comitement'' (what percentage of a user's stream is hate) or implicitness (e.g., the user tries to tread the fine line without violating the platform rules). For example, consider two users $u$ and $v$ posting utterances  \#3--\#8  in Table \ref{tab:hate_texts}. However, imagine that these are the only posts published by $u$, while $v$ published hundreds of other posts, non of which is hateful. Intuitively, one may argue that $v$'s questionable posts are diluted, compared to $u$'s, thus we should require a higher level of confidence in order to assign him the `hate monger' label (and compare to a user $v'$ posting the more explicit \#1 \& \#2 along with hundreds of non-hateful posts).

In order to address these variations we use distributional aggregation: instead of counting the user's hateful posts ($\Theta_F$), we look at the distribution of the user's posts as a $k$-dimensional vector where $k$ determines the number of bins used to approximate the probability density function.

Specifically, we consider two distributional spaces -- bins and quantiles. 
%Given that a user $u$ has $t_1,...,t_n$ posts ($T^u$), and assuming the distribution is represented in a $k$-dimensional vector, 
In the bin-based representation the [0,1] range is divided to $k$ equal bins, each bin (entry in the vector) holds the number of posts with the corresponding hate score assigned by $\theta(t)$. In the quantile-based representation the $k$ bins are unique for each user $u$ as they are defined over the range  $[min (\theta(t))$, $max(\theta(t)]$ for $t \in T^u$.

We therefore want to optimize
\vspace{-.2cm}
\begin{align}
    \Theta_D(u) = \sum_{i=1}^k w_i \cdot \sigma(B_i(u))
    \label{eq:distributional}
\end{align}

where $B_i(u)$ denotes the number of utterances for which $\theta(t)$ falls in the $i^{th}$ bin and $\sigma$ is the softmax function. 

\paragraph{Multimodal Aggregation} Finally, the different aggragation methods can be used together in a multimodal manner -- combining the hate-score distribution of the user utterances and the hate levels in the user's ego network. Using $\Theta^b_D$ and $\Theta^q_D$ to denote the bin-based and quantile-based representations we can formulate the combined model as
\vspace{-.2cm}
\begin{align}
    \Theta(u) = \Theta^b_D(u) + \Theta^q_D + \Theta^R
    \label{eq:multimodal}
\end{align}

%That is, having ten bins ($k=10$) and $\{t'_1, ...t'_m\} \subseteq T^x$ posts are assigned $0.2 \geq M(t'_i) \leq 0.3$, we have $\overrightarrow{x}_2=m$. 
%The quantile based representation the span of hate-score a user is assigned to her posts determines the bins, that is, if a the hate scores of a specific user $x$ fall between $\delta \geq 0$ (min) and $\delta' \leq 1$ (max) then the $j^{th}$ bin holds the number of posts with scores  in the range $[j \cdot \frac{\delta' - \delta}{k}, (j+1) \cdot \frac{\delta' - \delta}{k} )$.

\subsection{Social-aware Baselines}
\label{subsec:baselines}
In this section we briefly describe five strong baseline algorithms we use for comparison. These five algorithms, all leverage the structure of the social network and have proved useful in an array of node classification tasks, including the detection of hate speech and hateful users.

\paragraph{DeGroot's Diffusion} The DeGroot's model is a simple yet strong framework to classify nodes in a network through belief propagation. The model achieved good results in the detection of hateful users on Gab and Parler \cite{ribeiro2018characterizing,israeli2022free}. 

\paragraph{Graph Neural Networks (GNNs)}
\citet{das2021you} explored several GNN methods for detecting hateful users on Gab and Twitter. The network embeddings incorporate nodal features, e.g., textual representations of the user's utterances in learning the node and graph embeddings. In this work we use the following four algorithms:
\vspace{-.2cm}
\begin{enumerate}
    \item {\bf GCN}: Graph Convolutional Networks \cite{kipf2016semi}  uses localized approximation in learning node embeddings through convolutional layers.
    \vspace{-.2cm}
    \item {\bf GAT}: Graph Attention Network learn nodal representations by combining features of the nodes in the ego network of the focal node, setting their importance via attention layers.
    \vspace{-.2cm}
    \item {\bf GraphSAGE}: The Graph Sample and Aggregate \cite{hamilton2018inductive} learns nodal embedings by sampling a predefined number of nodes from the ego network of a focal node $u$, then applying attention layers in order to assign the importance of each node in the aggregative representation. 
    \vspace{-.2cm}
    \item {\bf{AGNN}}: The Attention-based Graph Neural Network \cite{thekumparampil2018attentionbased} uses attention layers to learn a dynamic and adaptive local summary of the neighborhood of each focal node. 
\end{enumerate} 

\vspace{-.2cm}

\section{Datasets and Annotation}
\label{sec:data}

\begin{table*}
    \centering
    {\small %defining the text as small
        \begin{tabular}{c@{\quad}|c@{\quad}|c@{\quad}c@{\quad}||c@{\quad}c@{\quad}|c@{\quad}c@{\quad}}
          \multicolumn{1}{c}{} & \multicolumn{1}{c}{} & \multicolumn{2}{c}{Raw Data} & \multicolumn{4}{c}{Annotated Data}\\[2pt]
          {Dataset} & {Source} & {\#Posts} & {\#Users} & {\#Posts} & {\% Hate } & {\#Users} & {\% Hate}\\[2pt]
          \hline\rule{0pt}{12pt}
            Echo (Twitter) & \citet{arviv2021sa} & 18M & 7.07K & 4630 & 8.2\% & 1000 & 15.4\%\\[2pt]
          \hline\rule{0pt}{12pt}
            Gab & \citet{alma9926823946704361} & 22M & 336.7K & 5000 & 5.1\% & 1000 & 24.8\%\\[2pt]
          \hline\rule{0pt}{12pt}
          Parler & This work & $183M^\dagger$ & $4.08M^\dagger$ & 8262$^\ddagger$ & 32.9\%$^\ddagger$ & 890 & 25.4\% \\[2pt]
          \hline
        \end{tabular}
    }
    \caption{Datasets statistics. The raw Parler data (marked $\dagger$) were shared by \cite{aliapoulios2021early} and the post-level annotations ($\ddagger$) were shared by \cite{israeli2022free}. User-level annotations are shared as part of this paper.}
    \label{table:table_datasets2}
\end{table*}

We evaluate our aggregative approaches over three very different datasets: Twitter-Echo, Gab, and Parler. The number of posts, users, and label breakdown for each dataset are provided in Table \ref{table:table_datasets2}. The remainder of this section provides further details regarding the datasets and our annotation process.

\paragraph{Echo (Twitter)}
\label{sec:data_echo}
The triple parentheses, or triple brackets, also known as the (((echo))), is an antisemitic symbol that is used to highlight the names of individuals of Jewish background (e.g., actress and comedian Amy Schumer, see utterance \#2 in Table \ref{tab:hate_texts}), organizations owned by Jewish people (e.g., Ben \& Jerry's), or organizations accused of promoting “Jewish globalist values”' (e.g., the International Monetary Fund).
The Echo dataset curated by \citet{arviv2021sa} contains over 18M English tweets posted by $\sim$7K echo users between May and June 2016. Annotations are provided at the tweet and the user level. An important feature of this dataset is that all users have utterances containing the echo symbol, although some users use it in a non-hateful manner, e.g., to symbolize a hug. This ambiguous nature of the symbol makes hate detection challenging. 

\paragraph{Gab}
\label{sec:data_gab}
Gab, launched on August 2016, was created as an alternative to Twitter, positioning itself as putting ``people and free speech first'', welcoming users suspended from other social networks. 
Gab posts (called \emph{gabs}) are limited to 300 characters, and users can repost, quote or reply to  previously created gabs. Gab permits pornographic and obscene content, as long as it is labeled \emph{NSFW} (`not safe for work').%Previous work finds the majority of Gab users to be Caucasians-conservatives-males \cite{lima2018inside}. 

The raw Gab dataset was introduced by \citet{zannettou2018gab}. It was collected using Gab’s API with the snowball methodology. More specifically, the researchers obtained data for the most popular users as returned by Gab’s API and iteratively collected data from all of their followers and their followees. They collected three types of information: basic details about Gab accounts (including username, score, and date of account creation); all the posts for each Gab user in the dataset; and all the followers and followees of each user, which allow the reconstruction of a social network. Overall, this dataset contains 22.1M posts from 336.8K users, posted between August 2016 and January 2018.
\citet{alma9926823946704361} shared an annotated dataset based on the aforementioned above, containing ~60K labeled posts and 1K labeled users.

\paragraph{Parler}
\label{sec:data_parler}
Alluding to the French verb `to speak', Parler was launched on August 2018.\footnote{On April 2023 the platform was acquired by Starboard and was taken offline to ``undergo a strategic assessment'' (Starboard announcement on Parler's landing page \url{https://parler.com/}, accessed: 5/8/2023). The platform was relaunched in February 2024 announcing it is ``breaking free from the constraints of conventional platforms'' (accessed: 6/5/2024).} 
The platform branded itself as ``The World's Town Square''  a place to \emph{``Speak freely and express yourself openly, without fear of being ``deplatformed'' for your views''}\footnote{Parler branding on its landing page (accessed: 3/10/2022)}.

Parler users post texts (called \emph{parlays}) of up to 1000 characters. Users can reply to parlays and to previous replies. 
%Parler has a voting system designated for ranking content -- following a simple upvote/downvote mechanism. Voting allows users to influence the order in which replies are displayed, much like the Reddit scoring order. 
Parler supports a reposting mechanism similar to Twitter's retweets (called `echos', not to confuse with the (((echo))) hate symbol, see above).
Parler's official guidelines\footnote{https://parler.com/documents/guidelines.pdf (accessed: 4/17/2022)} explicitly allowed ``trolling'' and ``not-safe-for-work'' (NSFW) content, include only three ``principles'' prohibiting ``unlawful acts'', citing ``Obvious examples include: child sexual abuse material, content posted by or on behalf of terrorist organizations, intellectual property theft''.

\citet{aliapoulios2021early} presented a dataset of tens of millions of Parler messages. \citet{israeli2022free} used this dataset to introduce an annotated dataset for hate speech (post level). Their 10K dataset consists of 3224 posts (32.8\%) labeled as hateful and 6053 (59.8\%) as non-hateful. However, a Parler \emph{user-based} annotated dataset has yet to be introduced. Hence, as part of this research, we create the first annotated dataset of Parler users.

\paragraph{Annotation of Parler Users} The Parler dataset presented by \citet{aliapoulios2021early} consists of $\sim$4M users. We focus on a subset of users matching the following criteria: (i) The account exists for at least six months; (ii) The user showed some activity (posted at least 30 posts); (iii) The primary language of the user is English. These three rules left us with a subset of users, denoted $U^*$, from which we sampled users for annotation. Following the protocol used by  \citet{ribeiro2018characterizing} and \citet{israeli2022free}, we used stratified subsampling mitigate bias (most users and vast majority of posts are not hateful). 

Annotation was done by 94 senior year Data Science B.Sc students for bonus course credit. Annotators were introduced to Parler and were given explicit instructions about the annotation task. The annotation process involved rating each account on a 1--5 scale (non-hateful -- extremely hateful). We ensured that each user is annotated by three annotators. The full annotation guidelined and further details regarding the annotation protocol are available in Appendix \ref{annotation_guidelines}.

\begin{table*}[ht!]
%\normalsize
    \centering
    \small
    \begin{adjustbox}{width=\textwidth}
        \begin{tabular}{c|c|c|c|c|c}
             & Method & Precision & Recall & F1 & ROC AUC\\
            \hline \hline\rule{0pt}{12pt}
            \multirow{8}{*}{\rotatebox[origin=c]{90}{Echo}}
            & DeGroot's Diffusion & $0.472 \pm 0.389$ & $0.255 \pm 0.261$ & $0.320 \pm 0.310$ & $0.610 \pm 0.122$\\
            & GCN & $0.443 \pm 0.118$ & $0.914 \pm 0.096$ & $0.585 \pm 0.104$ & $0.797 \pm 0.102$\\
            & GraphSAGE & $0.629 \pm 0.124$ & \bm{$0.950 \pm 0.041$} & $0.752 \pm 0.088$ & $0.944 \pm 0.024$\\
            & GAT & $0.548 \pm 0.131$ & $0.653 \pm 0.241$ & $0.574 \pm 0.123$ & $0.772 \pm 0.091$\\
            & AGNN & $0.759 \pm 0.072$ & $0.914 \pm 0.054$ & $0.826 \pm 0.035$ & $0.963 \pm 0.032$\\
            & Fixed-Threshold & $0.654 \pm 0.063$ & $0.627 \pm 0.095$ & $0.633 \pm 0.040$ & $0.836 \pm 0.040$\\
            & Relational Aggregation & \bm{$0.820 \pm 0.060$} & $0.834 \pm 0.055$ & $0.825 \pm 0.042$ & $0.956 \pm 0.013$\\
            & Distributional (bins) & $0.772 \pm 0.045$ & $0.871 \pm 0.058$ & $0.817 \pm 0.042$ & $0.944 \pm 0.019$\\
            & Distributional (quantiles) & $0.747 \pm 0.064$ & $0.899 \pm 0.047$ & $0.815 \pm 0.053$ & $0.942 \pm 0.018$\\
            & Distributional (bins+quantiles) & $0.757 \pm 0.058$ & $0.885 \pm 0.052$ & $0.815 \pm 0.049$ & $0.946 \pm 0.021$\\
            & Multimodal (relational+bins+quantiles) & $0.781 \pm 0.023$ & $0.899 \pm 0.048$ & \bm{$0.836 \pm 0.028$} & \bm{$0.961 \pm 0.011$}\\
            \hline\hline\rule{0pt}{12pt} 
            \multirow{8}{*}{\rotatebox[origin=c]{90}{Gab}}
            & DeGroot's Diffusion & $0.314 \pm 0.001$ & \bm{$0.777 \pm 0.000$} & $0.447 \pm 0.001$ & $0.604 \pm 0.000$\\
            & GCN & $0.241 \pm 0.109$ & $0.678 \pm 0.428$ & $0.334 \pm 0.169$ & $0.594 \pm 0.059$\\
            & GraphSAGE & $0.317 \pm 0.066$ & $0.559 \pm 0.206$ & $0.388 \pm 0.077$ & $0.582 \pm 0.077$\\
            & GAT & $0.194 \pm 0.115$ & $0.457 \pm 0.387$ & $0.264 \pm 0.174$ & $0.501 \pm 0.060$\\
            & AGNN & $0.340 \pm 0.031$ & $0.600 \pm 0.190$ & $0.423 \pm 0.030$ & $0.679 \pm 0.024$\\
            & Fixed-Threshold & \bm{$0.497 \pm 0.078$} & $0.351 \pm 0.066$ & $0.411 \pm 0.070$ & $0.722 \pm 0.039$\\
            & Relational Aggregation & $0.408 \pm 0.061$ & $0.437 \pm 0.086$ & $0.419 \pm 0.063$ & $0.675 \pm 0.043$\\
            & Distributional (bins) & $0.461 \pm 0.034$ & $0.649 \pm 0.044$ & $0.538 \pm 0.024$ & $0.763 \pm 0.016$\\
            & Distributional (quantiles) & $0.429 \pm 0.027$ & $0.702 \pm 0.056$ & $0.532 \pm 0.033$ & $0.770 \pm 0.018$\\
            & Distributional (bins+quantiles) & $0.435 \pm 0.026$ & $0.714 \pm 0.043$ & $0.540 \pm 0.029$ & $0.769 \pm 0.016$\\
            & Multimodal (relational+bins+quantiles) & $0.452 \pm 0.027$ & $0.702 \pm 0.042$ & \bm{$0.550 \pm 0.032$} & \bm{$0.777 \pm 0.020$}\\
            \hline\hline\rule{0pt}{12pt}  
            \multirow{8}{*}{\rotatebox[origin=c]{90}{Parler}}
            & DeGroot's Diffusion & $0.395 \pm 0.221$ & $0.441 \pm 0.247$ & $0.417 \pm 0.233$ & $0.644 \pm 0.081$\\
            & GCN & $0.284 \pm 0.054$ & \bm{$0.760 \pm 0.404$} & $0.348 \pm 0.130$ & $0.644 \pm 0.145$\\
            & GraphSAGE & $0.309 \pm 0.092$ & $0.649 \pm 0.189$ & $0.394 \pm 0.028$ & $0.497 \pm 0.061$\\
            & GAT & $0.379 \pm 0.051$ & $0.731 \pm 0.164$ & $0.488 \pm 0.013$ & \bm{$0.746 \pm 0.044$}\\
            & AGNN & $0.369 \pm 0.081$ & $0.552 \pm 0.255$ & $0.416 \pm 0.082$ & $0.667 \pm 0.069$\\
            & Fixed-Threshold & $0.470 \pm 0.050$& $0.369 \pm 0.040$& $0.412 \pm 0.035$& $0.693 \pm 0.026$\\
            & Relational Aggregation & \bm{$0.519 \pm 0.073$} & $0.509 \pm 0.082$ & \bm{$0.513 \pm 0.074$} & $0.730 \pm 0.051$\\
            & Distributional (bins) & $0.284 \pm 0.037$ & $0.500 \pm 0.075$ & $0.362 \pm 0.049$ & $0.575 \pm 0.024$\\
            & Distributional (quantiles) & $0.324 \pm 0.018$ & $0.734 \pm 0.057$ & $0.449 \pm 0.025$ & $0.611 \pm 0.040$\\
            & Distributional (bins+quantiles) & $0.324 \pm 0.021$ & $0.738 \pm 0.051$ & $0.450 \pm 0.027$ & $0.618 \pm 0.034$\\
            & Multimodal (relational+bins+quantiles) & $0.370 \pm 0.023$ & $0.680 \pm 0.082$ & $0.478 \pm 0.035$ & $0.699 \pm 0.044$\\
        \end{tabular}
    \end{adjustbox}
    \caption{5-Fold CV results on the test sets of Echo, Gab, and Parler datasets using the best (F1-score-wise) configuration. %Results marked with $\ast$ are statistically significant with $\alpha=0.05$
    }
    \label{tab:user_level_results}
\end{table*}

\section{Results and Analysis}
\label{sec:user_level_results}

\subsection{Experimental Settings}
\label{subsec:user_level_experiment}

For the utterance-level classification, we fine-tune DistilBERT  \cite{Sanh2019DistilBERTAD} on each datasets. We used a batch size of 32, a maximum number of epochs of 20 and a validation split of 0.2. We also used an early stopping callback with a patience of 5 epochs on the validation loss.

For the user-level classification, we use 5-Fold cross-validation for all of the methods. For each dataset, we considered the largest (weakly) connected component.
Specifically, for the GNN methods we used the same experimental settings as proposed by \citet{das2021you}. Appendix \ref{ap:networks} provides further details regarding the full networks, the number of connected components and the statistics regarding the largest connected component in each dataset.  

For the DeGroot's method, we followed the protocol used by \citet{israeli2022free}. 
%ran it for 10 iterations with a fraction of 5\% as a seed for hate-mongering users (nodes) with an initial belief of 1 (i.e. hate-mongers), and an initial belief of 0 for the rest of the nodes.
%In table \ref{tab:user_level_results} we report the best mean results of all the folds using each method for all evaluated datasets. The parameters used to achieve these results are reported in section \ref{sec:user_level_results}.
%While our proposed methods differ in the data preparation phase, some steps are common to all methods. For all of the methods we first need to retrieve the post-level predictions of all posts for each user we want to predict. For the relational aggregation method that uses the user's ego network, we also retrieved the post-level predictions of each user's followers and followees. 

\subsection{Results}

\paragraph{Utterance Level Prediction} All aggregative models require an initial step of classification of individual utterances. While the focus of this paper is the classification on the \emph{user} level, we report the results on the utterance level in order to highlight the challenge and the benefit of the aggregative methods. 

\begin{figure}[ht!]
    \centering
    \includegraphics[width=.5\textwidth]{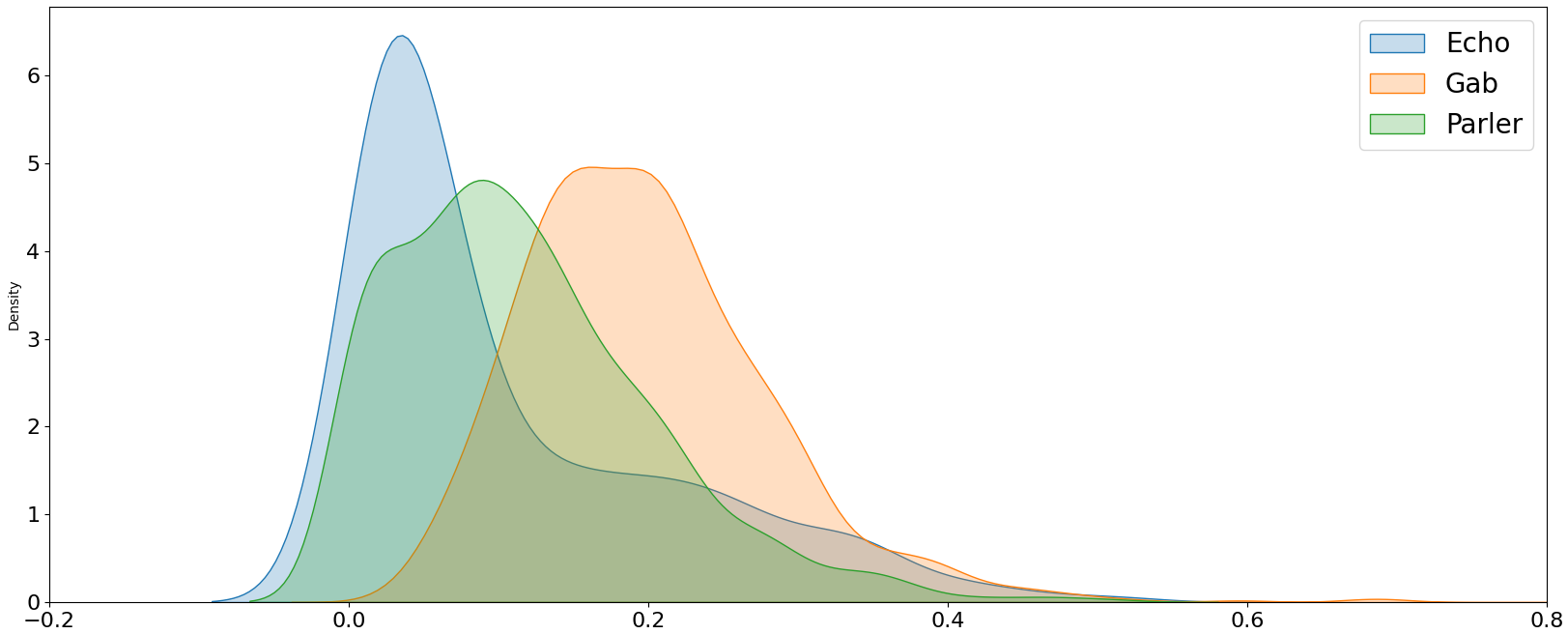}
    \caption{The Density of the mean utterance scores per user for each Dataset}
    \label{fig:post_mean_probas_per_user_kde}
\end{figure}

Results for each dataset are presented in Table \ref{tab:post_level_module_metrics}. Note the very low F-score achieved ob the Gab dataset and compare to the significant improvement achieved by shifting to the user level (below). The density of the mean hate score per user is presented in Figure \ref{fig:post_mean_probas_per_user_kde}, highlighting the differences between the datasets. 

\begin{table}[ht!]
    \centering
    \begin{tabular}{l|c|c|c}
        \hline
        \rule{0pt}{12pt}Dataset & Precision & Recall & F1 Score \\
        \hline\rule{0pt}{12pt}
        Echo & 0.4122 & 0.8026 & 0.5446 \\
        Gab & 0.2057 & 0.5472 & 0.2990 \\
        Parler & 0.6316 & 0.8177 & 0.7127 \\
        \hline
    \end{tabular}
    \caption{Performance metrics of the utterance-level model for Echo, Gab, and Parler Datasets}
    \label{tab:post_level_module_metrics}
\end{table}

\paragraph{User level results} Detailed results of each method over each of the three datasets are presented in Table \ref{tab:user_level_results}. 

Looking at the F-score achieved by the different methods we observe that our aggregative approach consistently outperform the baseline algorithms. While the multimodal aggregation yields the best results in the Echo and Gab datasets, it is ranked third on the Parler dataset, with the relational aggregation ranked first. We speculate that this is may be attributed to the size and the unique characteristics of the Parler network (see Appendix \ref{ap:networks}). A thorough analysis of the social networks is out of the scope of this paper.   

Breaking the multimodal aggregative model to separate aggregative models (Relational, Distributional-bins, Distributional-quantiles and bins and quantiles combined) shows that in most cases all these approaches are competitive with the top performing models and outperform the strong baselines. 

Interestingly, looking at the results of the relational method, we observe that the importance of the different components differ across datasets: best F1-score was achieved using $\alpha=0.608$, $\beta=0.776$, $\gamma=1.467$ for the Echo dataset; $\alpha=0.776$, $\beta=0.085$, $\gamma=0.108$ for the Gab dataset and $\alpha=0.239$, $\beta=0.254$, $\gamma=0.24$ for Parler. This result highlights the importance of the network structure and the dynamics and norms of each platform. 
This is also evident from the distribution of the mean utterance score in each platform.

\section{Conclusion}
\label{sec:conclusions}
We proposed a robust and efficient multimodal aggregative  method, combining text and social context through relational and distributional aggregations. We demonstrated the benefits of this approach for the task of hate speech and hatemonger detection over three unique and very different datasets from three social platforms: X (Twitter), Gab and Parler. 
%We evaluated our methods on both a large and unique corpus constructed around an ambiguous antisemitic meme and two other datasets based on platforms known for their free speech agenda and demonstrated their ability to correctly detect hate mongers across different social platforms. In addition, we showed how analyzing both the content and the network features of the users significantly improves the ability to detect hate mongers, and provides insights on the promotion of hate.

Future work takes three trajectories: (i) Improving the integration of the different modalities, and (ii) Better understanding the unique characteristics of each of the platforms and the datasets and the ways they impact the results, and (iii) Reapplication of the aggregative methods to improve classification of evasive utterances and emerging forms of coded hate speech. 

\newpage
\section{Limitations and Ethical Considerations} 

\subsection{Limitations} 
This work has a number of limitations, two of which are listed below (and other are referred to in the directions for future work):

(i) All three aggregation procedures depend on the basic classifier $\theta(t)$. An evasive user that is careful with his words, using only coded language or consistent gas-lighting may not be identified at all if $\theta(t)<\tau^T$ for all $t \in T^u$ for a user $u$.

(ii) We use the fixed threshold $\tau^U$ as a strong baseline, however, a more comprehensive comparison to other existing methods (e.g., diffusion-based and GNNs) should be considered in future work.

\subsection{Ethical Considerations} The detection of hate mongers on social platforms and the detoxification of the discourse are of great importance. However, wrongly labeling a user, may have severe emotional and legal implications. In this work, we explore computational approaches that can be used to study the phenomena at scale. As a practical policy, user labeling should be done carefully, ideally involving a `man-in-the-loop' procedure.

%In terms of future research, we recommend a broader examination of networks that includes both major social networks like Facebook and YouTube, as well as niche networks such as specific subreddits on Reddit and 4chan. Additionally, the analysis should incorporate a temporal analysis of user activity and network formation, while also improving the detection of implicit hate by combining the proposed methods. To further improve the accuracy of our analysis, we recommend using the output of user-level methods to reclassify posts. This approach allows us to refine our understanding of the content and context of posts, and can lead to more accurate identification of hate speech and related phenomena.
%Finally, we suggest exploring the use of graph-based algorithms such as Graph Convolution Networks\cite{https://doi.org/10.48550/arxiv.1609.02907} and Graph Attention Networks\cite{velickovic2017graph}, which are specifically designed to leverage the social semantics of each user. By utilizing these methods, we can gain a more comprehensive understanding of how hate speech and related phenomena are communicated and perpetuated across various networks.

\bibliography{hate_bib}
%\bibliographystyle{aaai22.bst}

%\newpage
\clearpage
\appendix
\section{Prompts and Generated Responses}
\label{sec:generated_predictions}
\texttt{Which of the following utterances may be perceived as hate speech?
\begin{enumerate}
    \item Good morning boys and girls!
    \item Muslim terrorists did this and Obama brought them here to run our government. Gotta get rid of Obammy, Omar, and all these other fucking Muslims.
    \item Is there a more debased creature in our modern pop-cultural cesspool than (((Amy Schumer)))? Probably not.
    \item We must secure the existence of our people and a future for white children!.
    \item If white people are going to continue existing - we must support our race.
    \item When the youngest white generation becomes adult they will be 5\% of the world population, down from 1/3 in 1900.
    \item Did you know that most dog breeds were bred by whites? We’re awesome and they’re awesome!
    \item White lives matter!
    \item Blue lives matter!
    \item Black lives matter!
\end{enumerate}
}
\paragraph{GPT-4} 
The utterances that may be perceived as hate speech are those that explicitly or implicitly promote hate, discrimination, or hostility towards particular groups based on their ethnicity, religion, or other inherent characteristics. Based on the examples you've provided, here's an analysis of which might be considered hate speech:
\begin{enumerate}
    \item[2.] \textbf{"Muslim terrorists did this and Obama brought them here to run our government. Gotta get rid of Obammy, Omar, and all these other fucking Muslims"} - This statement is clearly hate speech. It uses derogatory language targeting Muslims and makes unfounded, harmful generalizations about an entire religious group.
    \item[3.] \textbf{"Is there a more debased creature in our modern pop-cultural cesspool than (((Amy Schumer)))? Probably not."} - This statement might be perceived as hate speech, particularly because the use of triple parentheses is often a coded way to identify someone as Jewish in an anti-Semitic context.
    \item[4.] \textbf{"We must secure the existence of our people and a future for white children!"} - This phrase is widely recognized as a white supremacist slogan, known as the "Fourteen Words." It is used to promote the idea of white racial superiority and is considered hate speech.
    \item[5.] \textbf{"If white people are going to continue existing - we must support our race"} - This could be interpreted as promoting racial supremacy or separatism, which might be viewed as hate speech depending on the context in which it is used.
\end{enumerate}
The other statements, while potentially controversial or promoting particular views, do not explicitly convey hate against a group based on inherent characteristics in the same way that the identified statements do.
\\
\\
\paragraph{Gemini 1.5 Pro}
\textbf{Analysis of Potentially Hateful Utterances:}
Here's a breakdown of the provided utterances and their potential to be perceived as hate speech:

\begin{enumerate}
    \item \textbf{Good morning boys and girls!}
    \begin{itemize}
        \item This is a simple greeting and does not contain any hateful content.
    \end{itemize}

    \item \textbf{Muslim terrorists did this and Obama brought them here to run our government. Gotta get rid of Obammy, Omar and all these other fucking Muslims.}
    \begin{itemize}
        \item This statement employs harmful stereotypes against Muslims, associating them with terrorism and expressing a desire to remove them from the government. This is a clear example of hate speech.
    \end{itemize}

    \item \textbf{Is there a more debased creature in our modern pop-cultural cesspool than (((Amy Schumer)))? Probably not.}
    \begin{itemize}
        \item The use of triple parentheses around a name is a known anti-semitic symbol online. This statement is likely intended to target Amy Schumer for her Jewish heritage and is therefore hate speech.
    \end{itemize}

    \item \textbf{We must secure the existence of our people and a future for white children!}
    \item \textbf{If white people are going to continue existing - we must support our race}
    \item \textbf{When the youngest white generation becomes adult they will be 5\% of the world population, down from 1/3 in 1900}
    \begin{itemize}
        \item These three statements, while not explicitly hateful, promote a white supremacist ideology. They focus on the preservation of white people and imply a fear of their decline, potentially inciting hostility towards other racial groups.
    \end{itemize}

    \item \textbf{Did you know that most dog breeds were bred by whites? We’re awesome and they’re awesome!}
    \begin{itemize}
        \item This statement, while expressing pride in an accomplishment associated with white people, does not directly attack or demean any other group. It is unlikely to be considered hate speech.
    \end{itemize}

    \item \textbf{White lives matter!}
    \item \textbf{Blue lives matter!}
    \item \textbf{Black lives matter!}
    \begin{itemize}
        \item These statements, on the surface, simply affirm the value of different groups of people. However, depending on the context and intent, they can be interpreted in different ways.
        \item \textit{White lives matter} is often used as a response to the "Black Lives Matter" movement, implying a denial of the systemic racism faced by Black people. In such cases, it can be considered hate speech.
        \item \textit{Blue lives matter} expresses support for law enforcement. While not inherently hateful, it can be used to dismiss or oppose calls for police accountability and racial justice.
        \item \textit{Black lives matter} aims to highlight the systemic racism and violence faced by Black communities. It is a call for justice and equality, not a statement of superiority or hatred towards other groups.
    \end{itemize}
\end{enumerate}

\textit{It's crucial to consider the context and intent behind these statements to determine if they constitute hate speech.}

% \begin{figure*}[ht!]
%     \centering
%     \includegraphics[width=1.\textwidth]{resources/images/results_horizontal_2.png}
%     \caption{Results of all methods}
%     \label{fig:results}
% \end{figure*}

% \begin{figure*}[ht!]
%     \centering
%     \includegraphics[width=1.\textwidth]{resources/images/results_horizontal.png}
%     \caption{Results of all methods}
%     \label{fig:results_horizontal}
% \end{figure*}

\begin{table*}[ht!]
    \centering
    \begin{tabular}{c|c|c|c|c|c}
         Dataset & Threshold & Precision & Recall & F1 & ROC AUC \\
        \hline \hline
        \multirow{4}{*}{Echo}
         & 1 & 0.264 & 1.000 & 0.417 & 0.506 \\
         & 3 & 0.266 & 1.000 & 0.420 & 0.511 \\
         & 10 & 0.277 & 1.000 & 0.434 & 0.538 \\
         & 50 & 0.358 & 0.950 & 0.520 & 0.673 \\
         & 100 & 0.433 & 0.820	& 0.567	& 0.721 \\
        \hline \hline
        \multirow{4}{*}{Gab}
         & 1 & 0.252 & 1.000 & 0.402 & 0.506 \\
         & 3 & 0.257 & 1.000 & 0.409 & 0.519 \\
         & 10 & 0.281 & 0.984 & 0.437 & 0.573 \\
         & 50 & 0.380 & 0.559 & 0.452 & 0.628 \\
         & 100 & 0.437 &	0.351 &	0.389 &	0.600 \\
        \hline \hline
        \multirow{4}{*}{Parler}
         & 1 & 0.308 & 0.883 & 0.457 & 0.608 \\
         & 3 & 0.341 & 0.824 & 0.482 & 0.644 \\
         & 10 & 0.348 & 0.689 & 0.462 & 0.627 \\
         & 50 & 0.457 & 0.387 & 0.420 & 0.616 \\
         & 100 & 0.607 &	0.293 &	0.395 & 0.615 \\
    \end{tabular}
    \caption{Performance metrics for Echo, Gab, and Parler datasets}
    \label{tab:performance_metrics_combined}
\end{table*}

\section{Networks Statistics}
\label{ap:networks}
Tables \ref{tab:table_datasets3} and  \ref{tab:table_datasets4} provide more information on each of the datasets and justifies the focus on the largest connected component in each. 

\begin{table*}
    \centering
    {\small %defining the text as small
        \begin{tabular}{c@{\quad}|c@{\quad}c@{\quad}c@{\quad}|c@{\quad}c@{\quad}}
          % \multicolumn{1}{c}{} & \multicolumn{5}{c}{Raw Data}\\[2pt]
          {Dataset} & {\#Posts} & {\#Users} & {\#Edges} & {\#Connected Components} & {\#Singletons}\\[2pt]
          \hline\rule{0pt}{12pt}
            Echo (Twitter) & 18M & 7.07K & 21.4K & 9075 & 2919\\[2pt]
          \hline\rule{0pt}{12pt}
            Gab & 19.42M & 61.36K & 2.63M & 40K & 10.13K\\[2pt]
          \hline\rule{0pt}{12pt}
            Parler (this work) & 115M & 3.08M & 11.14M & 5.45M & 2.43M \\[2pt]
          \hline
        \end{tabular}
    }
    \caption{Datasets statistics for the full network, excluding users without any posts.}
    \label{tab:table_datasets3}
\end{table*}

\begin{table*}
    \centering
    {\small %defining the text as small
        \begin{tabular}{c@{\quad}|c@{\quad}c@{\quad}c@{\quad}|c@{\quad}c@{\quad}|c@{\quad}c@{\quad}}
          \multicolumn{1}{c}{} & \multicolumn{5}{c|}{Raw Data} & \multicolumn{2}{c}{Annotated Data}\\[2pt]
          {Dataset} & {\#Posts} & {\#Users} & {\#Edges} & {Clustering Coefficient} & {Optimal Gamma} & {\#Users} & {\% Hate}\\[2pt]
          \hline\rule{0pt}{12pt}
            Echo (Twitter) & 9.8M & 3.7K & 20.7K & 0.19 & 2.8 & 532 & 26.1\%\\[2pt]
          \hline\rule{0pt}{12pt}
            Gab & 19.28M & 51.2K & 2.47M & 0.402 & 4.06 & 982 & 24.5\%\\[2pt]
          \hline\rule{0pt}{12pt}
            Parler (this work) & 60.7M & 643K & 11.4M & 0.224 & 2.14 & 881 & 25.2\%\\[2pt]
          \hline
        \end{tabular}
    }
    \caption{Datasets statistics for the largest (weakly) connected component. We treated the graphs as undirected in order to calculate the Clustering Coefficient and the Optimal Gamma (assuming a Power-law distribution)}
    \label{tab:table_datasets4}
\end{table*}

\section{Annotation Guidelines}
\label{annotation_guidelines}
Each annotator was assigned 50 Parler accounts, ensuring that each account is being annotated by three students. For each account, annotators were given the user name, self-description of the user's account, and 30 posts published by the user. The 30 posts are sampled out of the following: (i) 15 posts with the highest hate speech prediction values\footnote{The prediction is according to the BERT model introduced by \citet{israeli2022free}}; (ii) Its 10 most recent posts; (ii) Five random posts published by the account. The annotators showed a reasonable agreement level with an average Cohen's Kappa of 0.36.  The full guidelines for the Parler dataset annotation process can be found here
\url{https://mega.nz/file/D0pjBDDR#9iMK0uC72w41BjSj_eDCmpbaJ6YGU8sZ8g0dvDv0pcU}
% [$https://mega.nz/file/D0pjBDDR#9iMK0uC72w41BjSj_eDCmpbaJ6YGU8sZ8g0dvDv0pcU$].

\section{Responsible NLP Research Checklist}
\begin{enumerate}
    \item[A1] Did you describe the limitations of your work? \textbf{Yes}
    \item[A2] Did you discuss any potential risks of your work? \textbf{Yes}
    \item[A3] Do the abstract and introduction summarize the paper’s main claims? \textbf{Yes}
    \item[B] Did you use or create scientific artifacts? \textbf{Yes}
    \item[B1] Did you cite the creators of artifacts you used? \textbf{Yes}
    \item[B2] Did you discuss the license or terms for use and / or distribution of any artifacts? \textbf{Yes, we discuess the limitations and ethical considerations of our work}
    \item[B3] Did you discuss if your use of existing artifact(s) was consistent with their intended use, provided that it was specified? For the artifacts you create, do you specify intended use and whether that is compatible with the original access conditions (in particular, derivatives of data accessed for research purposes should not be used outside of research contexts)? \textbf{Yes}
    \item[B4] Did you discuss the steps taken to check whether the data that was collected / used contains any information that names or uniquely identifies individual people or offensive content, and the steps taken to protect / anonymize it? \textbf{No, since our work involves harmful or offensive content.}
    \item[B5] Did you provide documentation of the artifacts, e.g., coverage of domains, languages, and linguistic phenomena, demographic groups represented, etc.? \textbf{Yes}
    \item[B6] Did you report relevant statistics like the number of examples, details of train / test / dev splits, etc. for the data that you used / created? \textbf{Yes}
    \item[C] Did you run computational experiments? \textbf{Yes}
    \item[C1] Did you report the number of parameters in the models used, the total computational budget (e.g., GPU hours), and computing infrastructure used? \textbf{Yes}
    \item[C2] Did you discuss the experimental setup, including hyperparameter search and best-found hyperparameter values?] \textbf{Yes}
    \item[C3] Did you report descriptive statistics about your results (e.g., error bars around results, summary statistics from sets of experiments), and is it transparent whether you are reporting the max, mean, etc. or just a single run? \textbf{Yes}
    \item[C4] If you used existing packages (e.g., for preprocessing, for normalization, or for evaluation), did you report the implementation, model, and parameter settings used (e.g., NLTK, Spacy, ROUGE, etc.)? \textbf{Yes}
    \item[D] Did you use human annotators (e.g., crowdworkers) or research with human participants? \textbf{Yes}
    \item[D1] Did you report the full text of instructions given to participants, including e.g., screenshots, disclaimers of any risks to participants or annotators, etc.? \textbf{Yes}
    \item[D2] Did you report the full text of instructions given to participants, including e.g., screenshots, disclaimers of any risks to participants or annotators, etc.? \textbf{Yes}
    \item[D3] Did you discuss whether and how consent was obtained from people whose data you’re using/curating \textbf{Yes}
    \item[D4] Was the data collection protocol approved (or determined exempt) by an ethics review board? \textbf{N/A}
    \item[D5] Did you report the basic demographic and geographic characteristics of the annotator population that is the source of the data? \textbf{Yes}
    \item[E] Did you use AI assistants (e.g., ChatGPT, Copilot) in your research, coding, or writing? \textbf{Yes}
    \textbf[E1] Did you include information about your use of AI assistants? \textbf{We used LLMs as classifiers, see Table \ref{tab:hate_texts} and references through the paper.}

\end{enumerate}

\end{document}